\documentclass[sigconf]{acmart}

\usepackage{booktabs} 
\usepackage{graphicx} 

\setcopyright{rightsretained}

\acmConference[Deep Learning' 18]{Penn State Deep Learning Fall 2018}{Fall 2018}{State College, PA, USA}
\acmYear{2018}
\copyrightyear{2018}

\begin{document}
\title{End to End Video Segmentation for Driving : Lane Detection For Autonomous Car}

\author{Tejas Mahale Chaoran Chen and Wenhui Zhang * equal contribution}
\orcid{1234-5678-9012}
\affiliation{%
  \institution{Penn State University}
  \city{State College}
  \state{Pennsylvania}
  \postcode{16803}
}
\email{wenhui@gwmail.gwu.edu}

\renewcommand{\shortauthors}{WZ et al.}

\begin{abstract}
Safety and decline of road traffic accidents remain important issues of autonomous driving. Statistics show that unintended lane departure is a leading cause of worldwide motor vehicle collisions, making lane detection the most promising and challenge task for self-driving. Today, numerous groups are combining deep learning techniques with computer vision problems to solve self-driving problems. In this paper, a Global Convolution Networks (GCN) model is used to address both classification and localization issues for semantic segmentation of lane. We are using color-based segmentation is presented and the usability of the model is evaluated. A residual-based boundary refinement and Adam optimization is also used to achieve state-of-art performance. As normal cars could not afford GPUs on the car, and training session for a particular road could be shared by several cars. We  propose a framework to get it work in real world. We build a real time video transfer system to get video from the car, get the model trained in edge server (which is equipped with GPUs), and send the trained model back to the car. 

\end{abstract}


\maketitle

\section{Introduction}
Statistics show that unintended lane departure is a leading cause of worldwide vehicle collisions, which leads to substantial financial costs to both society and the individuals involved [1]. To reduce the number of traffic accidents and to improve safety, research on Driver Assistance System (DAS) have been conducted worldwide for many years and even expanded to autonomous cars. 
Autonomous car is always equipped with intelligent vehicle (IV) system, and lane detection is an important information for this system.  In this system, some tasks including road following, keeping within the correct lane, maintaining a safe distance between vehicles, controlling the speed of a vehicle according to traffic conditions and road conditions, moving across lanes in order to overtake vehicles and avoid obstacles, searching for the correct and shortest route to a destination have to be completed [2]. The premise of all the tasks are high related to the lanes. Only when we know the accurate lanes will we finish above tasks better. Furthermore, The distance between the lane boundaries and obstacle is also important because it makes sure that the vehicle is in a safe distance from another vehicle or obstacles to avoid any possible collisions[2]. To obtain the information of the distance to the lane boundaries, we need to know the lane boundaries first. That is why lane detection is an important part of intelligent vehicle system.\\
Lane detection mainly focus on locating both the center line and the edge of each lane in a road image. lane markings help differentiate the lanes from other characteristic objects on the road such as other vehicles, pedestrian, and animals running into the road. But there are some challenges we have to face. Firstly, does markings really mark the correct lane? As we know, there are various road conditions and environment conditions that lanes maybe not visible in a image. Sometimes the quality of the image can also be the reason. Apart from that, road splitting or merging and the interference from roadside objects or shadows could be another problem which will worsen the detection.\\
However, the great variety of road environments necessitates the use of complex vision algorithms that not only requires expensive hardware to implement but also relies on many adjustable parameters that are typically determined from experience [3]. With the development of the deep learning techniques, numerous groups have applied a variety of deep learning techniques to computer vision problems [4]. \\
In this paper, we gather original data from Carla simulator In Implementation Details section, we do some pre-processing work to clean the data, which is a preparation for using a image segmentation techniques to deal with the lane detection problem. The method is based on color to differentiate different objects. Then in Experiments and Evaluation section, we conduct a Global Convolution Networks (GCN) model using color-based segmentation and use Adam as an optimizer to optimize the results. After obtaining the results(using test data to get training image), we compare the image we get with the actual image to evaluate the model. 

In summary, we highlight our potential contributions below:
\begin{itemize}
	\item This report introduces a fine grained architecture of GCN model;
	\item We implements a GCN network with optimization method of Residual-based boundary refinement and Adam optimization;
	\item This paper discusses and conducts performance diagnostics of lane detection tasks under various weather conditions.
	\item We proposed a framework to get the model trained and tested in real world.  
\end{itemize}

The rest of this report is structured as follows. Section \ref{sec:related-works} gives out overview for related works.
Section \ref{sec:formular} explains and gives out an overview on algorithms and math background of our model.
In Section \ref{sec:implementation} , we present our detailed implementation for data pre-processing pipeline and data augmentation.
In Section \ref{sec:evaluation} , we present our proposal for evaluating performance analysis for our lane detection model.
It also explains performance of our model on train, validation and test data.

\section{Related Work}
\label{sec:related-works}
\subsection{Image Segmentation Techniques}
Image segmentation is the operation of partitioning an
image into a collection of connected sets of pixels.
Main methods of region segmentation includes, region growing, clustering and split and merge. Region growing techniques start with one pixel of a 
potential region and try to grow it by adding adjacent
pixels till the pixels being compared are too far from similar. The first pixel selected can be just the first unlabeled pixel in the image or a set of seed pixels can be chosen from the image. Usually a statistical test is used to decide which pixels can be added to a region. [6] Clustering methods includes  K-means Clustering and Variants, Isodata Clustering, Histogram-Based Clustering and Recursive Variant and Global-Theoretic Clustering. [7] In some image sets, lines, curves, and circular arcs are more useful than regions or helpful in addition to regions. Thus people use lines and arcs segmentation as basis of image segmentation. Basic idea is looking for a neighborhood with strong signs of change, and detect them as edge. 
Edge detectors are based on differential operators.
Differential operators attempt to approximate the gradient at a pixel via masks. And then threshold the gradient to select the edge pixels. One widely used edge detector used in auto driving lane detection in earlier days is Canny Edge Detector (CED). CED firstly smooth the image with a Gaussian filter, then compute gradient magnitude and direction at each pixel of the smoothed image. It zeros out any pixel response smaller than the two neighboring pixels on either side of it,  along the direction of the gradient, and track high-magnitude contours. Afterwards it keeps only pixels along these contours, so weak little segments go away. [8]

These classical image segmentation methods are good, however they do not work well with raining or snowing road situations. It is hard for these classical image segmentation methods to handle various scenarios. Thus higher order image segmentation methods, such as deep learning's convolution neural network (CNN) is a must in auto driving. 

\subsection{Structured Deep Learning}

Classical machine learning on structured datasets methods includes kernel-based methods and graph-based regularization techniques [1, 2]. However these machine learning requires a lot of feature engineering before certain tasks such as image classification. Furthermore, most of the time these features require domain knowledge, creativity and a lot of trial and error. Structured Deep Learning (SDL)  is a fast, no domain knowledge requiring, and high performing machine learning method.

In the last couple of years, a number of papers re-visited machine learning on structured datasets, like graphs. Some works has done in generalizing neural networks to work on arbitrarily structured graphs [4, 5].  Some of them achieve promising results in domains that have previously been dominated by classical machine learning methods mentioned above.
Kipf  [9] came up with a graph convolutions method, which is good for image segmentation. Global convolutions are generalization of convolutions, and easiest to define in spectral domain. General Fourier transform scales poorly with size of data so we need relaxations. In this paper they use first order approximation in Fourier-domain to obtain an efficient linear-time graph-CNNs. We adapt this approach in our paper. We illustrate here what this first-order approximation amounts to on a 2D lattice one would normally use for image processing, where actual spatial convolutions are easy to compute
in this application the modelling power of the proposed graph conv-networks is severely impoverished, due to the first-order and other approximations made. It uses concept of a convolution filter for image pixels or a linear array of signals.

\subsection{Optimization Techniques on Gradients}

Optimization algorithms used to accelerate convergence includes stochastic gradient methods and stochastic momentum methods. Stochastic gradient methods can generally be written 
\begin{equation}\label{eq:desc}
w_{k+1} = w_k - \alpha_k\, \tilde \nabla f(w_k),
\end{equation}  
where $\tilde \nabla f(w_k) := \nabla f(w_k; x_{i_k})$ is the gradient of some loss function $f$ computed on a batch of data $x_{i_k}$.  

Stochastic momentum methods have been used to accelerate training. These methods could be written as
\begin{align}\label{eq:gen-mom}
w_{k+1} &= w_k - \alpha_k \, \tilde \nabla f(w_k + \gamma_k (w_k-w_{k-1})) + \beta_k(w_{k} - w_{k-1}).
\end{align}
Sequence of iterates~\eqref{eq:gen-mom} includes Polyak's heavy-ball method (HB) with $\gamma_k = 0$, and Nesterov's Accelerated Gradient method (NAG) with $\gamma_k = \beta_k$. 

However there are some exceptions as well, such as  \eqref{eq:desc} and \eqref{eq:gen-mom}. These are adaptive gradient and adaptive momentum methods. These methods construct the entire sequence of iterates, say 
$(w_1, \cdots, w_k)$ as a local distance measure. AdaGrad [10],  RMSProp [11], and Adam [12] can generally be written as 
\begin{align}\label{eq:adap-mom}
w_{k+1} &= w_k - \alpha_k \mathrm{H}_k^{-1}\tilde \nabla f' +\beta_k\mathrm{H}_k^{-1} \mathrm{H}_{k-1} (w_k - w_{k-1}),
\end{align} 
where $\tilde \nabla f'$ is $\tilde \nabla f(w_k + \gamma_k(w_k-w_{k-1}))$, and  $\mathrm{H}_k:= H(w_1, \cdots, w_k)$ is a positive definite matrix. $\mathrm{H}_k$ is a diagonal matrix. Its entries are defined as square roots of a linear combination of squares of past gradient components.  

In deep learning acceleration of convergence, specific settings of the parameters are stated in Table~\ref{table:update}. 
In this table, ${D}_k$ = $({g}_k\circ{g}_k)$ and 
${G}_k$ = ${H}_k\circ{H}_k$.

As we could see from Table~\ref{table:update}, adaptive methods change and adapt to geometry of the data. However, stochastic gradient descent and related variants use the $\ell_2$ geometry inherent. This is considered as equivalent to making$\mathrm{H}_k=\mathrm{I}$ in adaptive methods. Performance is defined as  loss function than the function $f$ used in training.  Thus, it seems like Adam is more favorable in our GCN project.

\begin{table}[h]
\centering
\tiny
\begin{tabular}{c|c|c|c|c|c|c}
& SGD & HB & NAG & AdaGrad & RMSProp & Adam  \\
\hline
$\mathrm{G}_k$ &  $\mathrm{I}$ &  $\mathrm{I}$&  $\mathrm{I}$&$\mathrm{G_{k-1}} +  \mathrm{D}_k$ &$\beta_2 \mathrm{G_{k-1}} + (1 - \beta_2) \mathrm{D}_k$& $\frac{\beta_2}{1- \beta_2^k} \mathrm{G_{k-1}} + \frac{(1 - \beta_2)}{1- \beta_2^k}\mathrm{D}_k$ \\
$\alpha_k$ & $\alpha$&$\alpha$&$\alpha$&$\alpha$&$\alpha$&$\alpha \frac{1-\beta_1}{1- \beta_1^k}$ \\
$\beta_k$ & 0   & $\beta$ & $\beta$& 0  &0 &$\frac{\beta_1(1 - \beta_1^{k-1})}{1 - \beta_1^k}$ \\
$\gamma$ & 0 & 0 & $\beta$ & 0 & 0 & 0 
\end{tabular}
\vspace{10pt}
\caption{Parameter Settings of Optimization Techniques 
used in Deep Learning. }
\label{table:update}
\end{table}

\section{Formulation}
\label{sec:formular}
Lane detection involves following steps. 
Firstly, compute the camera calibration matrix and distortion coefficients given a set of chessboard images.
Then, apply a distortion correction to raw images, and use color transforms, gradients, etc., and apply deep learning for classification analysis to create a threshold binary image. Afterwards, from a birds-eye view, apply a perspective transform to rectify binary image.
Then, detect lane pixels and fit to find the lane boundary, and determine the curvature of the lane and vehicle position with respect to center.
The last step is to warp the detected lane boundaries back onto the original image. Last but not least, put visual display of the lane boundaries and numerical estimation of lane curvature and vehicle position.

\begin{figure}[h]
\begin{minipage}[t]{0.7\linewidth} 
\centering    
\includegraphics[width=1\textwidth]{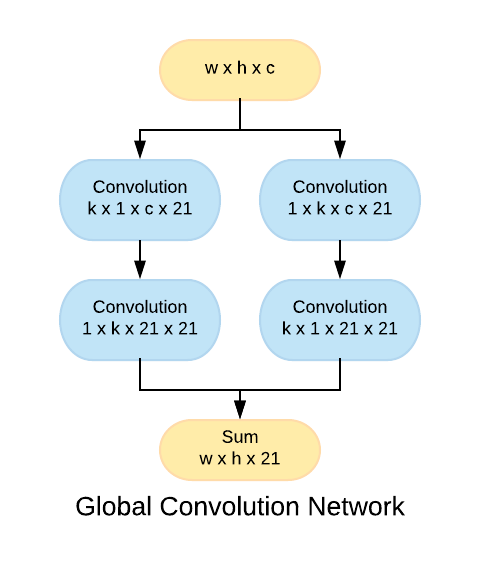}
\caption{Details of GCN}
\label{fig:GCN}
\end{minipage}
\hfill
\begin{minipage}[t]{0.7\linewidth}
\centering
\includegraphics[width=1\textwidth]{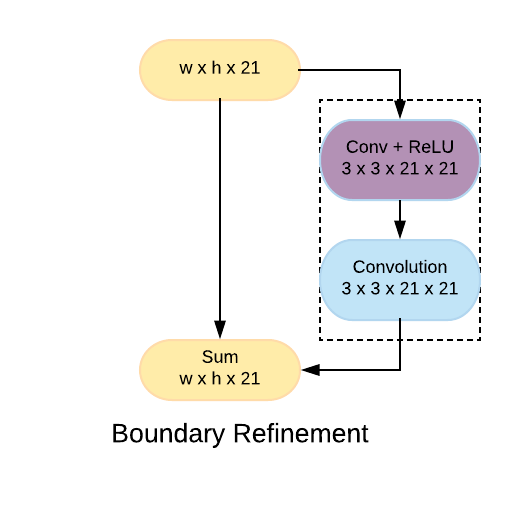}
\caption{Details of BR}
\label{fig:BR}
\end{minipage}
\end{figure}

In this paper, the GCN utilizes connectivity structure of graph as the filter to perform neighborhood mixing. Its architecture may be elegantly summarized as:
\[
H^{(l+1)}=\sigma(\hat{A}H^{(l)}W^{(l)}),
\]
where $\hat{A}$ is normalization of the graph adjacency matrix, $H^{(l)}$ is row-wise embedding of graph vertices in the $l$th layer, $W^{(l)}$ is a parameter matrix, and $\sigma$ is non-linearity.

in this paper we consider a building block defined as: $y= F(x, W_{i}) + x.$ Here $x$ and $y$ are the input and output vectors of a particular layer. The function $F(x, W_{i})$ represents learning residual mapping. $F=W_{2} * \sigma (W_{1}x)$ in which $\sigma$ denotes ReLU and the biases are omitted for simplifying notations. The operation $F+x$ is performed by a shortcut connection and element-wise addition. We adopt the second non-linearity after the addition.

\begin{figure*}[ht]
  \begin{center}
     \includegraphics[width=1\linewidth]{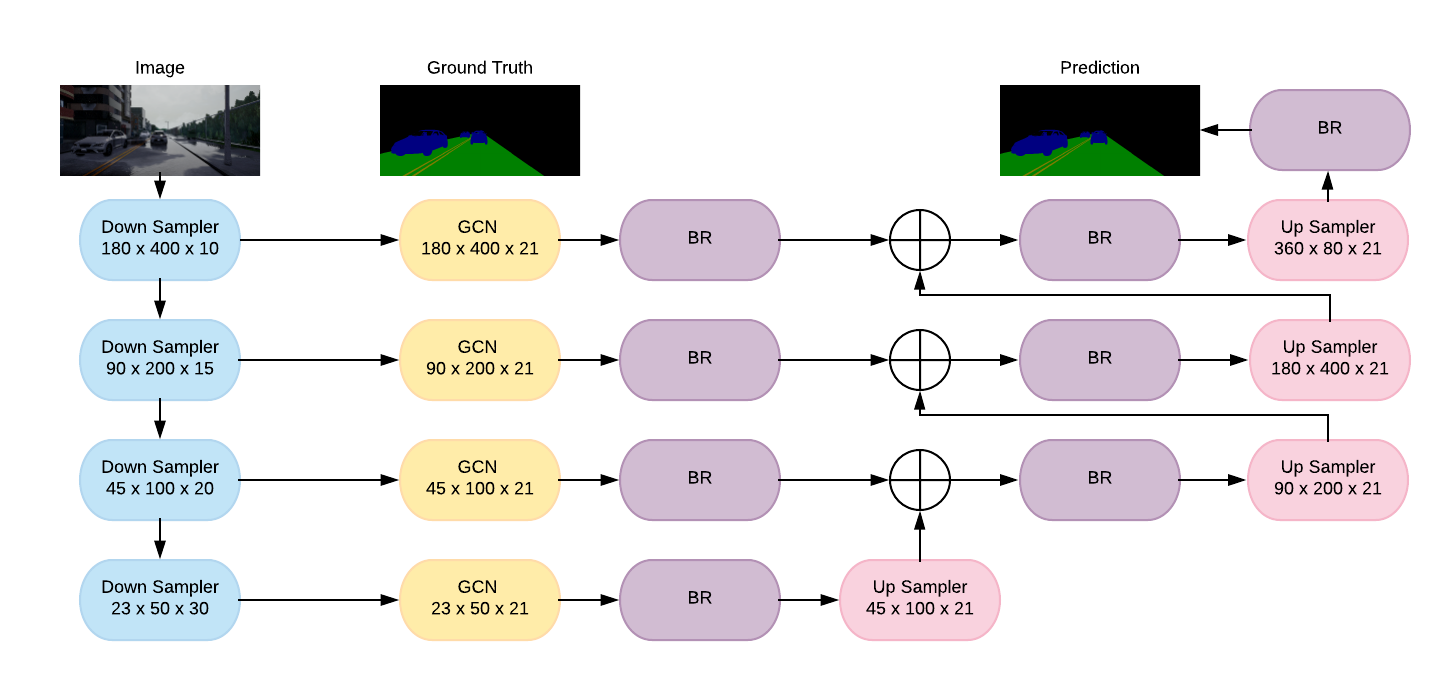}
  \end{center}
  \caption{ An overview of the whole pipeline.}
  \label{fig:whole-pipeline}
\end{figure*}

In this section, we first propose a novel GCN to address the contradictory aspects --- classification and localization in semantic segmentation. Then using GCN \ref{fig:GCN} we design a fully-convolutional framework for semantic segmentation task. Comparing with other many semantic segmentation models, our model has faster training speeds with smaller model size, while keeping similar accuracy.  In features extraction stage, we use the max pooling module following by conv-layers and an up-sampling layer. After these two modules, feature maps has same size as input image. When the enlargement factor is large, it brings noise and make boundary pixels of two areas difficult to classify. And that is why we add refinement module at the end of the model to refine the segmentation area \ref{fig:BR} In refinement module, we use combination of convolution layers and pooling layers, with a conv-net with ReLU, and then process it with convolution of size  3x3x21x21. 

In convolution neural network training, network weights are adjusted by loss function evaluation. As images vary from class to class, influence of each class to the loss is different. In each iteration, we calculate the weights based on current input batch. Weights are different in each iteration. 
The weights is calculated as in Eq.\ref{equ1}.
\begin{equation}
w_i = \left\{ 
             \begin{array}{lc}  
             1, & n_i = 0	\\
             \beta,	&	w_i < \beta\\	
             \dfrac{N}{2*c*n_i},	&	\beta < w_i < \alpha\\
             \alpha,	&	w_i > \alpha
             \end{array}  
\right.  
\label{equ1}
\end{equation}

Where, $w_i$ is weight of class $i$, $c$ is class number, and value of $i$ is from 0 to $c$.
$\beta$ and $\alpha$ are lower and upper threshold of $w_i$, we set threshold to avoid excessive weights differences. 
$N$ is the total pixel number of this batch, $n_i$ is the pixel number of class $i$, when $n_i$ = 0, it means that the class $i$ does not appear in this batch, we set the weight to 1. 
Because we need to increase the effect of small pixel number class on loss, so the smaller the $n_i$, the larger the $w_i$ is. $N$ and $c$ are constant, $w_i$ is just changed by $n_i$. 
When the $n_i$ is the average number, $w_i$ is calculated to be $\dfrac{1}{2}$, the multiplicative coefficient of $\dfrac{1}{2}$ is also used to decrease the $w_i$ of large pixel number of class.
The loss function is shown in Eq.\ref{equ2}, where $x_{ij}$, $y_{ij}$ are prediction class and label in pixel (i, j), $w$ is the loss weights.

\begin{equation}
LOSS = \sum_{i = 1}\sum_{j = 1}w\left \| x_{ij} - y_{ij} \right \|^2
\label{equ2}
\end{equation}

And the whole architecture could be seen as shown in Fig ~\ref{fig:whole-pipeline}.

\subsection{Encode - Decoder}

To decodify our depth first we get the int24.  $R + G*256 + B*256*256$. Then normalize it in the range $[0, 1]$. $Ans / ( 256*256*256 - 1 )$. And finally multiply for the units that we want to get. We have set the far plane at 1000 metres. Ans * far. Encoded stage includes encoding image to segmented image. Every pixel of label image is red channel class label. We use retained class labels with road and vehicle and converted them to green and blue respectively.
Encoder network consist of convolution layers, batch normalization and max pooling for down-sampling[13]. For classification problems, multiple convolution layers with decrease in size of kernel and increase in number of filters with each layer pooling is advised[14]. On other hand decoder block consists of strided convolution and up-sampler which will re-size image into original shape.
When we transfer the trained image back to color map to be project to videos, we just use $blue = value/(256*256)$, $ green = (value - blue*256*256)/256$ and $red = value - green*256 + blue*256*256$.

\subsection{GCN and BR}
There are two tasks in segmentation, classification and localization. Conical CNN architectures are good for classifications. In case of object segmentation, large kernel size plays important role in term of object localization[15]. But usage of large kernels is heavy on weights of model as kernel size increases, number of parameters also increases. To achieve global convolution (GCN) performance with less weights, GCN was divided into two one dimensional kernels. Effect of both kernels added at end to get actual GCN output.
Boundary refinement block(BR) is residual block which helps to get boundary structure of shape properly.

\subsection{Adam}
In this paper, we use Adam. Similar to momentum, it keeps exponentially decaying of past gradients $m_t$. It compute bias-corrected first and second moment estimates:

\begin{align}
\begin{split}
\hat{m}_t &= \frac{m_t}{1 - \beta^t_1}\\
\hat{v}_t &= \frac{v_t}{1 - \beta^t_2}
\end{split}
\end{align}

$m_t$ and $v_t$ are estimates of the first moment and the second moment of the gradients respectively. Parameters are updated using:
\begin{equation}
\theta_{t+1} = \theta_{t} - \frac{\eta}{\sqrt{\hat{v}_t} + \epsilon} \hat{m}_t
\end{equation}

In this paper, we use default values of $0.9$ for $\beta_1$, $0.999$ for $\beta_2$, and $10^{-8}$ for $\epsilon$.

\section{Implementation Details}
\label{sec:implementation}
\subsection{Dataset}
\begin{figure}[h]
\begin{minipage}[t]{0.4\linewidth} 
\centering    
\includegraphics[width=1\textwidth]{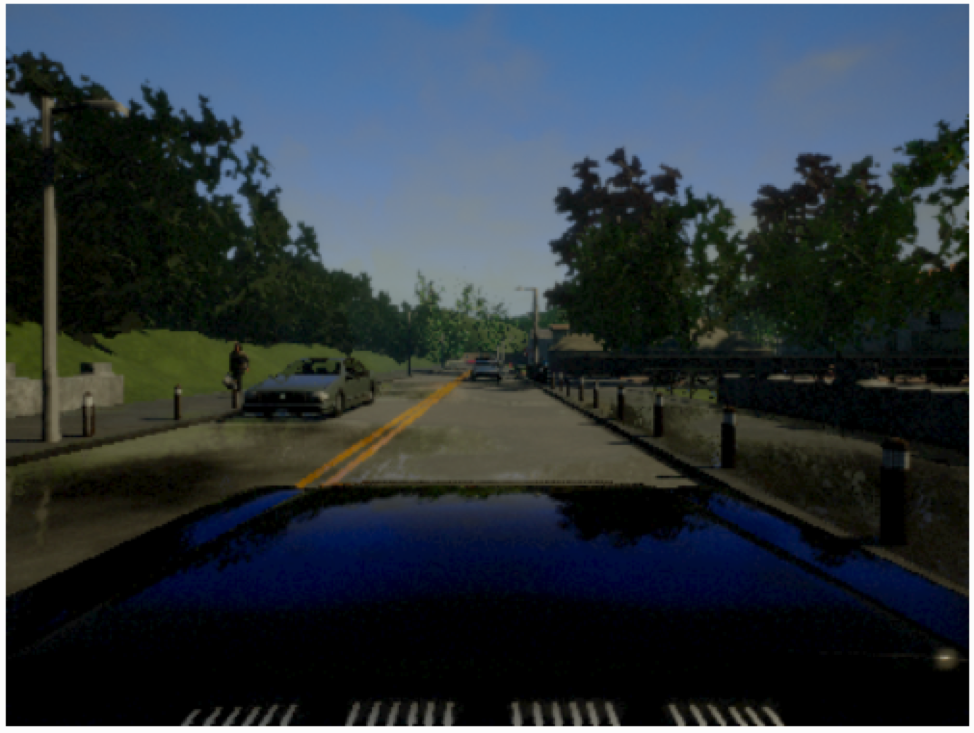}
\caption{original image}
\label{original image}
\end{minipage}
\hfill
\begin{minipage}[t]{0.4\linewidth}
\centering
\includegraphics[width=1\textwidth]{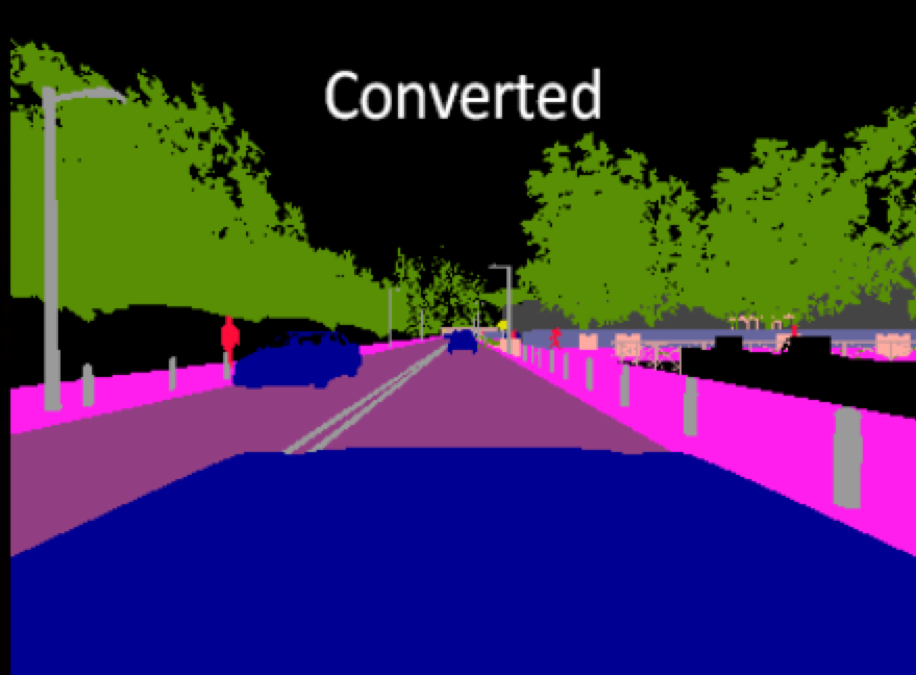}
\caption{converted image}
\label{converted image}
\end{minipage}
\end{figure}

First, we gathered our 3000 road images from a platform named Carla Simulator. It is an open-source simulator especially for autonomous driving research. It is a good tool since it provides various different images in a variety of circumstances including different types of  weather conditions and different time of day. Through driving a car in urban conditions, we could get train images(shown in Figure \ref{original image}) and corresponding encoded label images(shown in Figure \ref{converted image}). 3,000 images along with semantic labels with resolution 600 x 800 x 3 are collected as our dataset.


\subsection{Data Pre-processing}

\begin{figure}[h]
\begin{minipage}[t]{0.4\linewidth} 
\centering    
\includegraphics[width=1.0\textwidth]{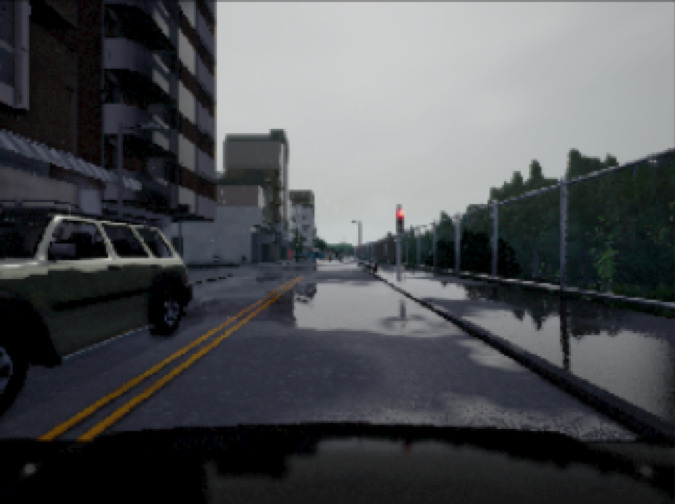}
\caption{600*800*3}
\label{600*800*3}
\end{minipage}
\hfill
\begin{minipage}[t]{0.4\linewidth}
\centering
\includegraphics[width=1.0\textwidth]{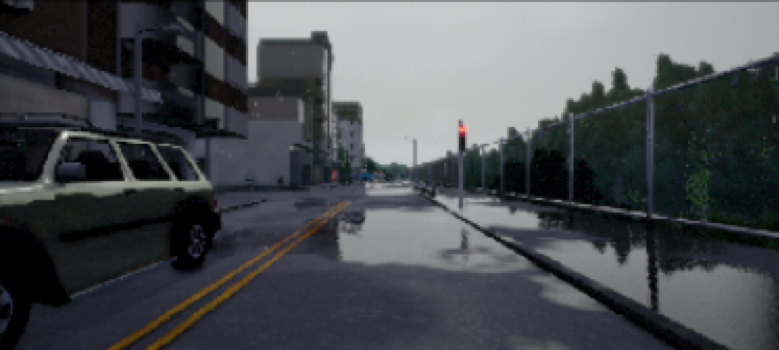}
\caption{360*800*3}
\label{360*800*3}
\end{minipage}
\end{figure}

To reduce the workload of computer and accelerate the running speed, car hood and sky portion are cropped in each image for 240*800 pixels(From Figure \ref{600*800*3} to Figure \ref{360*800*3}). That leads to a dataset consisting of 360*800 images of the red, green, and blue color channels.

\begin{figure}[h] 
	\centering  
	\includegraphics[width=0.4\linewidth]{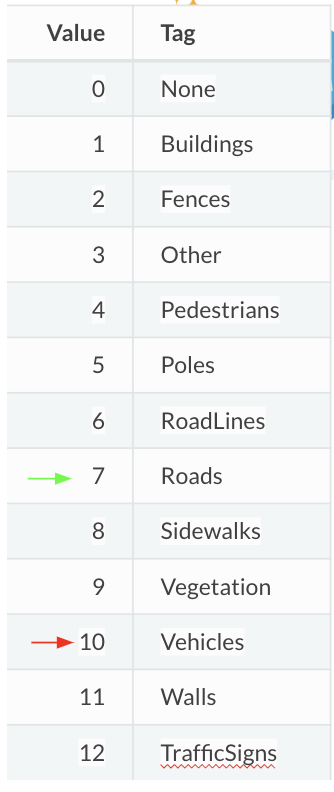}  
	\caption{Tags}  
	\label{fig:tags}   
\end{figure}

\begin{figure}[h] 
	\centering  
	\includegraphics[width=0.7\linewidth]{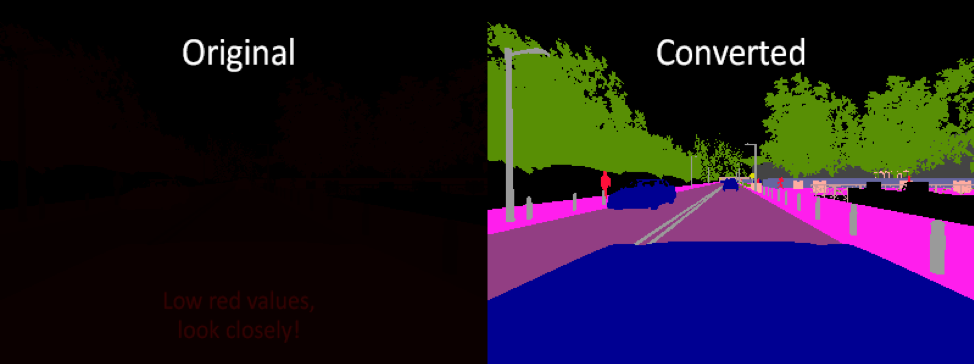}  
	\caption{encoded image to Segmented image(1)}  
	\label{fig:encoded image}   
\end{figure}

\begin{figure}[h] 
	\centering  
	\includegraphics[width=0.7\linewidth]{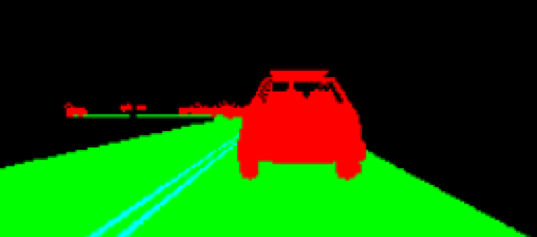}  
	\caption{encoded image to Segmented image(2)}  
	\label{fig:segmemted image}   
\end{figure}

The server provides an image with the tag information encoded in the red channel. A pixel with a red value of x displays an object with tag x. The following tags are displayed (shown in Figure \ref{fig:tags}). For training purposes, encoded image(shown in Figure \ref{fig:encoded image}) should be transformed to segmented image(shown in Figure \ref{fig:segmemted image}). We retained class labels with road and vehicle and converted them to green and blue respectively.

\subsection{Data Augmentation}

\begin{figure}[h]
\begin{minipage}[t]{0.4\linewidth} 
\centering    
\includegraphics[width=1.0\textwidth]{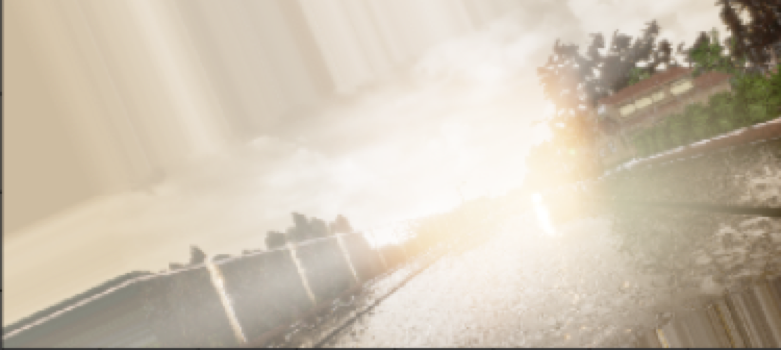}
\caption{Rotation}
\label{fig:rotation0}
\end{minipage}
\hfill
\begin{minipage}[t]{0.4\linewidth}
\centering
\includegraphics[width=1.0\textwidth]{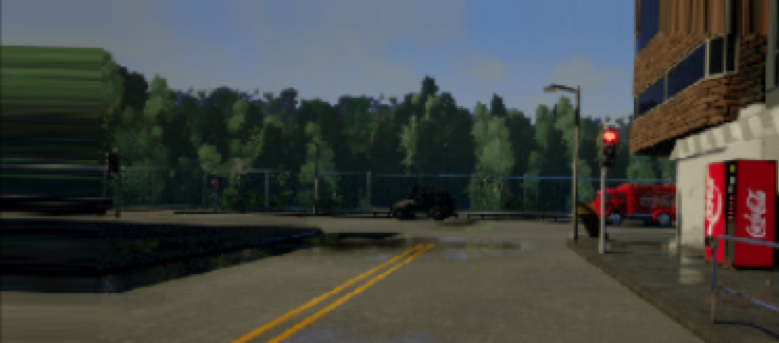}
\caption{Shifting}
\label{fig:shifting0}
\end{minipage}
\end{figure}

\begin{figure}[h]
\begin{minipage}[t]{0.4\linewidth} 
\centering    
\includegraphics[width=1.0\textwidth]{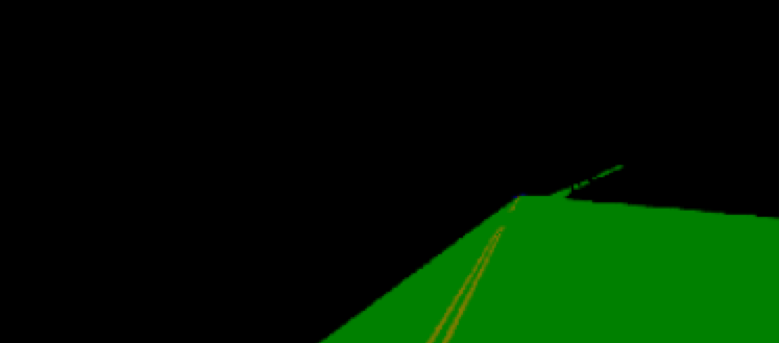}
\caption{Rotation}
\label{fig:rotation1}
\end{minipage}
\hfill
\begin{minipage}[t]{0.4\linewidth}
\centering
\includegraphics[width=1.0\textwidth]{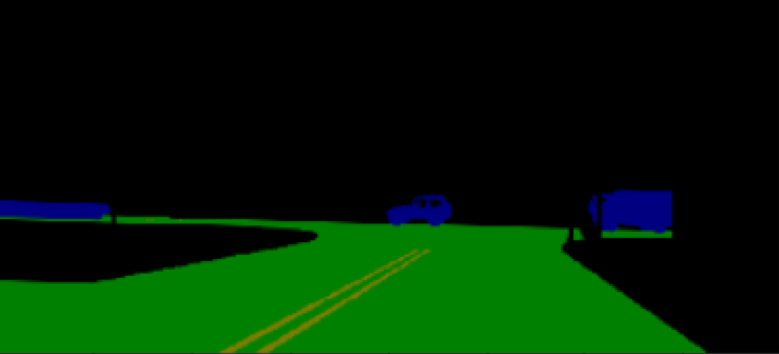}
\caption{shifting}
\label{fig:shifting1}
\end{minipage}
\end{figure}

To ensure the diversity of dataset, we made some data augmentation. As is shown in Figure \ref{fig:rotation0} and Figure \ref{fig:rotation1}, Rotation (0, 30 degree) like that is given to the original dataset to get another 3000 images, while in Figure \ref{fig:shifting0} and Figure \ref{fig:shifting1}, shifting (Width = (0, 0.2), height = (0, 0.1)) like that is given to the original dataset as well to get other 3000 images. In total, we obtained 9000 images dataset at present, which is of course a large amount of images and also make sure the diversity of the dataset. 

\section{Evaluation and Discussion}
\label{sec:evaluation}
\subsection{Metric}

In this paper, evaluation on both Global Convolution Network and  Boundary refinement are conducted. We evaluate our GCN based lane detection model in terms of minimum loss performance with help of minimum square error and mean average error.
Our experiment setup is composed of one class of server. 
For training and experiments performed on Tesla P100-PCIE gpu memoryclock rate 1.3285 with 11.95 GB video memory on Linux distribution.Conda environment is used with tensorflow gpu v1.10, Keras gpu v2.2.2, python version 3.5.

Our implementation for training is as follows. We re-size images with its shorter side randomly sampled in 600 x 800 x 3 for scale augmentation 360 x 800 x 3 resolution.
 The standard color augmentation mentioned in previous section is used. Batch normalization is adopted right after each convolution layer before activation. We initialize the weights as per Xavier initializer,  and train encoder nets from scratch. We
use Adam with a global-batch size of 64. The learning rate
starts from 0.001 and the models are trained for up to 40
iterations. We use a weight decay of 0.9 and a momentum of 0.999.

In order to judge whether a lane segmentation is successfully detected, we view lane markings as two classified value and calculate the  Minimum Square Error (MSE) and Mean Absolute Error (MAE) between the ground truth and the prediction. MSE is average of square of pixel to pixel difference between two images where as MAE is pixel to pixel absolute difference between images.

 MSE = $\frac{1}{N}\sum_{n=1} ^ {N} (Y_{true} - Y_{predicted})^ {2}$\\
 
 MAE = $\frac{1}{N}\sum_{n=1}^{N} |(Y_{true} - Y_{predicted})|$

\subsection{Hyper-parameter tuning}
Hyper-parameter tuning is important process of deep learning algorithms. In this project, we have implemented encoder-decoder architecture which has parameters like number of filters in each encoder layer, filter size, learning rate of optimizer, number of iterations to get minimum loss. \\
\begin{center}
\begin{tabular}{|l|r|}

	\hline
	Parameter & Value\\
	\hline 
    Optimizer & Adam\\
	\hline
    Learning rate & 0.001\\
	\hline
	Number of iterations & 40\\
	\hline
	Number of Filters per layer & 8, 16, 20, 32\\
	\hline
\end{tabular}
Table 5.1
\end{center}
As explained earlier, we implemented conical structure for encoder blocks which consist of convolution layer, batch normalization and max pooling to half the image resolution at every step along with gradual increase in number of filters. We trained our model for 5-10 iterations of various combination of hyper-parameter and observed the decay in mean square loss. Finally we came with hyper-parameter settings that give minimum loss at 40th iteration.

\subsection{Training and Validation Results}
We divided our image dataset into train, validation and test categories. We trained our model on train data and tuned hyper-parameter on validation data.

\begin{center}
Minimum Square Error (MSE)
\begin{tabular}{|l|c|r|}

	\hline
	Iterations & Training (MSE) & Validation (MSE) \\
	\hline
    0 Initial loss & 1592.1951 &  1443.7032 \\
	\hline
    10 Iterations & 192.1637 & 337.5560 \\
	\hline
	20 Iterations & 102.3213 & 119.4761 \\
	\hline
	30 Iterations & 74.0532 & 89.0769 \\
	\hline
	40 Iterations & 42.7771 & 61.4360 \\
	\hline
\end{tabular}
Table 5.2
\end{center}

\begin{center}
Mean Average Error (MAE)
\begin{tabular}{|l|c|r|}

	\hline
	Iterations & Training (MAE) & Validation (MAE) \\
	\hline
    0 Initial loss & 13.3800 &  13.3908 \\
	\hline
    10 Iterations & 3.1369 & 6.8221 \\
	\hline
	20 Iterations & 2.9696 & 3.7389 \\
	\hline
	30 Iterations & 2.0202 & 2.8999 \\
	\hline
	40 Iterations & 1.7011 & 2.4999 \\
	\hline
\end{tabular}\\
Table 5.3
\end{center}

After 40 iterations, train loss was further decreasing but on same time validation loss started increasing, so we had to stop training process. We can see final iteration values of MAE and MSE and conclude that we have got optimized parameters with minimum loss and less over-fitting.  

\begin{figure}[h]
\begin{minipage}[t]{0.4\linewidth}
\begin{center}    
\includegraphics[width=1.9\textwidth]{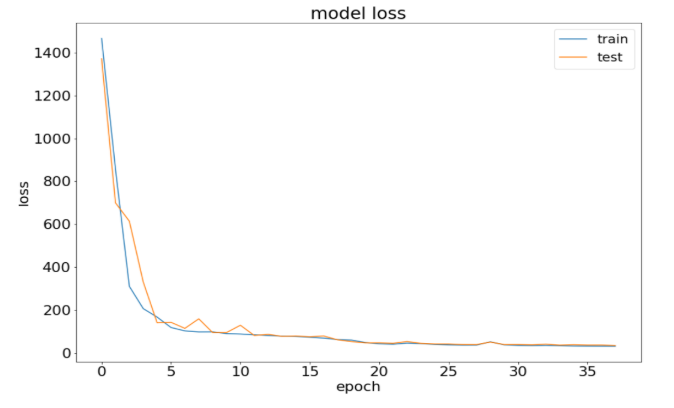}
\caption{MSE for Train and Validation }
\label{fig:1}
\end{center}
\end{minipage}
\end{figure}

\subsection{Test Results}
Once we concurred best hyper-parameters for minimum loss on given model with help of MSE and MAE values, we tested our model on reserved test dataset to get unbiased evaluation of our model.

\begin{center}

\begin{tabular}{|l|r|}

	\hline
	Metric & Test set evaluation \\
	\hline
    Minimum Square Error & 57.5875 \\
	\hline
    Mean Absolute Error &  2.2104 \\
	\hline
	
\end{tabular}\\
Table 5.4
\end{center}

From table 5.2 and 5.4, we can compare MSE values on train, validation and test set. There is not much difference in train and validation MSE. Also, validation and train values are almost in range. It implies that model is not facing over-fitting problem.

\begin{figure}[h]
\begin{minipage}[t]{0.4\linewidth} 
\centering    
\includegraphics[width=1.1\textwidth]{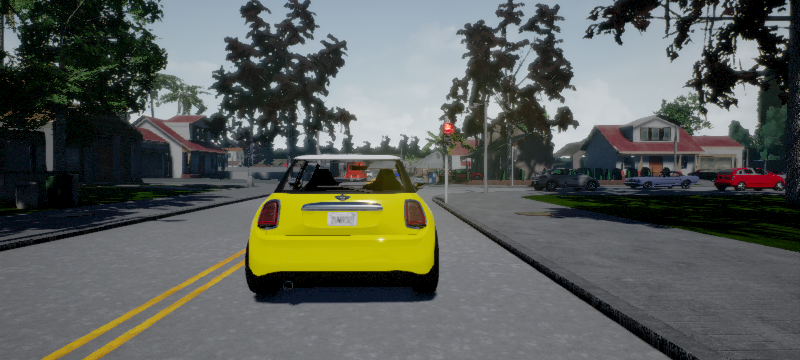}
\caption{Test image 1}
\label{fig:1}
\end{minipage}
\hfill
\begin{minipage}[t]{0.4\linewidth}
\centering
\includegraphics[width=1.1\textwidth]{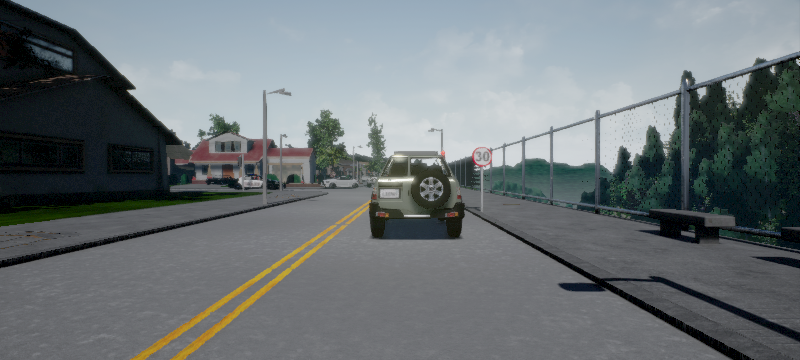}
\caption{Test Image 2}
\label{fig:t2}
\end{minipage}
\end{figure}

\begin{figure}[h]
\begin{minipage}[t]{0.4\linewidth} 
\centering    
\includegraphics[width=1.1\textwidth]{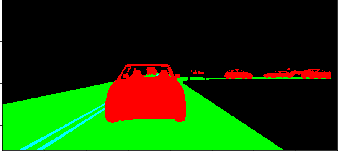}
\caption{Ground truth of image 1}
\label{fig:1}
\end{minipage}
\hfill
\begin{minipage}[t]{0.4\linewidth}
\centering
\includegraphics[width=1.1\textwidth]{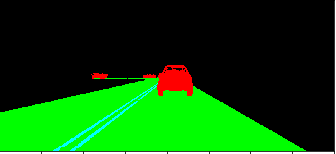}
\caption{Ground truth of Image 2}
\label{fig:g2}
\end{minipage}
\end{figure}

\begin{figure}[h]
\begin{minipage}[t]{0.4\linewidth} 
\centering    
\includegraphics[width=1.1\textwidth]{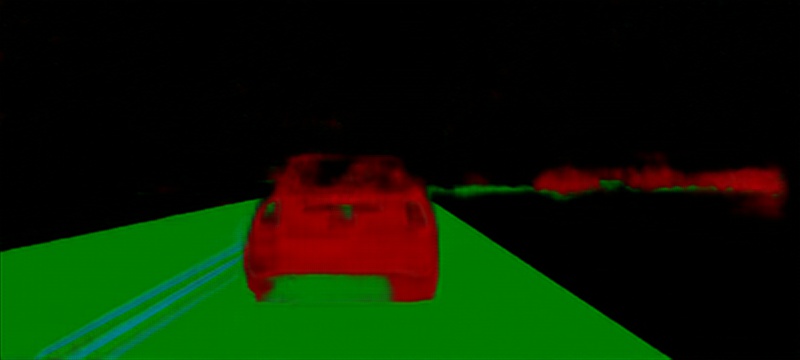}
\caption{Model Prediction image 1}
\label{fig:1}
MSE = 48.8590
\end{minipage}
\hfill
\begin{minipage}[t]{0.4\linewidth}
\centering
\includegraphics[width=1.1\textwidth]{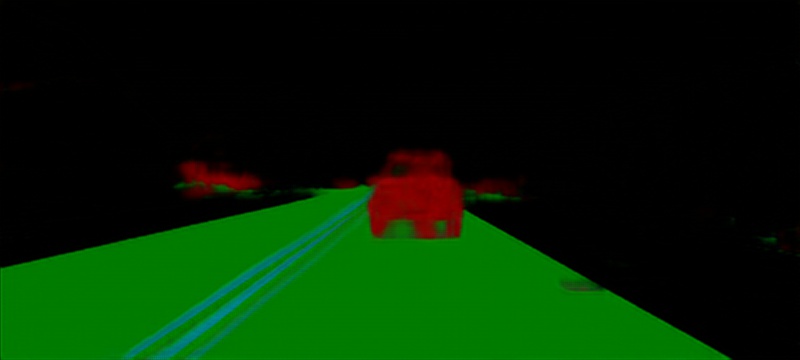}
\caption{Model Prediction image 2}
\label{fig:m2}
MSE = 63.3626
\end{minipage}
\end{figure}
We calculated minimum square error for test image1 and test image2 from test dataset. Both images were able to capture shape of vehicle as well as lane boundaries. From ground truth label image and model predicted image, we compared MSE score of both images.

\begin{table}[t]
\begin{center}
 \begin{tabular}{|l|c|}
    \hline
    Method & MSE  \\
    \hline
    FCN 8s & 65.3 \\
    \hline
    DPN  & 59.1 \\
    CRFasRNN  & 62.5 \\
    Scale invariant $CNN + CRF$ & 66.3 \\
    Dilation10  &  67.1 \\
    DeepLabv2-CRF  & 70.4 \\
    Adelaide\_context  & 71.6 \\
    LRR-4x  & 71.8   \\
    Large-Batch  & 76.9   \\
    \hline 
    \textbf{Our approach} & \textbf{57.5875}\\
    \hline
 \end{tabular}\\
 Table 5.5 Experimental results on test set
\end{center}

\end{table}

\section{Conclusion}
\label{sec:conclusion}
 In this paper, a Global Convolution Networks (GCN) model is used to address both classification and localization issues in lane detection. A residual-based boundary refinement and Adam optimization is also used to achieve 57.5875 performance. We also build a picamera on RaspberryPi and put it on a car, this will send real time video to our edge server. On edge server side, we will have training scripts running and sending trained GCN model back to the car. The code could be found here:
 
 $https://github.com/wenhuizhang/auto\_driving\_car$

\bibliographystyle{ACM-Reference-Format}

\bibliography{sample-bibliography}
\section{References}
[1] Narote, S. P., Bhujbal, P. N., Narote, A. S., \& Dhane, D. M. (2018). A review of recent advances in lane detection and departure warning system. Pattern Recognition, 73, 216-234.

[2] Chiu, K. Y., Lin, S. F. (2005, June). Lane detection using color-based segmentation. In Intelligent Vehicles Symposium, 2005. Proceedings. IEEE (pp. 706-711). 

[3] Yim, Y. U., Oh, S. Y. (2003). Three-feature based automatic lane detection algorithm (TFALDA) for autonomous driving. IEEE Transactions on Intelligent Transportation Systems, 4(4), 219-225.

[4]Huval, B., Wang, T., Tandon, S., Kiske, J., Song, W., Pazhayampallil, J., ... \& Mujica, F. (2015). An empirical evaluation of deep learning on highway driving. arXiv preprint arXiv:1504.01716.

[3] Martin Abadi et al. TensorFlow: Large-scale machine learning on heterogeneous systems, 2015.

[4] Bruna, J., Zaremba, W., Szlam, A., \& LeCun, Y. (2013). Spectral networks and locally connected networks on graphs. arXiv preprint arXiv:1312.6203.

[5] Henaff, M., Bruna, J., \& LeCun, Y. (2015). Deep convolutional networks on graph-structured data. arXiv preprint arXiv:1506.05163.

[6] Al-Hujazi, E., \& Sood, A. (1990). Range image segmentation combining edge-detection and region-growing techniques with applications sto robot bin-picking using vacuum gripper. IEEE transactions on systems, man, and cybernetics, 20(6), 1313-1325.

[7] Lucchese, L., \& Mitra, S. K. (2001). Colour image segmentation: a state-of-the-art survey. Proceedings-Indian National Science Academy Part A, 67(2), 207-222.

[8] Ding, L., \& Goshtasby, A. (2001). On the Canny edge detector. Pattern Recognition, 34(3), 721-725.

[9] Kipf, T. N., \& Welling, M. (2016). Semi-supervised classification with graph convolutional networks. 
arXiv preprint arXiv:1609.02907.

[10] Duchi, J., Hazan, E., \& Singer, Y. (2011). Adaptive subgradient methods for online learning and stochastic optimization. Journal of Machine Learning Research, 12(Jul), 2121-2159.

[11] Tieleman, T., \& Hinton, G. (2012). Rmsprop: Divide the gradient by a running average of its recent magnitude. coursera: Neural networks for machine learning. COURSERA Neural Networks Mach. Learn.

[12] Kingma, D. P., \& Ba, J. (2014). Adam: A method for stochastic optimization. arXiv preprint arXiv:1412.6980.

[13] “Autoencoders, Unsupervised Learning, and Deep Architectures”, Pierre Baldi, 2012

[14] "Very Deep Convolutional Networks for Large-Scale Image Recognition", Karen Simonyan, Andrew Zisserman, September 2014

[15] "Large Kernel Matters -- Improve Semantic Segmentation by Global Convolutional Network", Chao Peng, Xiangyu Zhang, Gang Yu, Guiming Luo, Jian Sun, March 17

\end{document}